\documentclass[letterpaper, 10 pt, conference]{ieeeconf}  
\pdfoutput=1

\IEEEoverridecommandlockouts                              

\usepackage{graphics} 
\usepackage{subfigure}
\usepackage{epsfig} 
\usepackage{amsmath} 
\usepackage[T1]{fontenc}
\usepackage{bm}
\usepackage{cite}
\usepackage{booktabs}
\usepackage{multirow}
\usepackage{diagbox}
\usepackage{threeparttable}
\usepackage{hyperref}
\hypersetup{
    colorlinks=false, 
    }
\hypersetup{hidelinks} 

\title{\LARGE \bf
Shape Control of Deformable Linear Objects with Offline and Online Learning of Local Linear Deformation Models
}

\author{Mingrui Yu, Hanzhong Zhong, and Xiang Li
\thanks{M. Yu, and X. Li are with the Department of Automation, Tsinghua University, China, and H. Zhong is with the School of Aerospace Engineering, Tsinghua University, China.
This work was supported in part by the Science and Technology Innovation 2030-Key Project under Grant 2021ZD0201404, in part by the Institute for Guo Qiang, Tsinghua University, and in part by the National Natural Science Foundation of China under Grant U21A20517. Corresponding author: Xiang Li (xiangli@tsinghua.edu.cn)}}

\begin{document}

\maketitle
\thispagestyle{empty}
\pagestyle{empty}

\begin{abstract}
The shape control of deformable linear objects (DLOs) is challenging, since it is difficult to obtain the deformation models. Previous studies often approximate the models in purely offline or online ways. In this paper, we propose a scheme for the shape control of DLOs, where the unknown model is estimated with both offline and online learning. The model is formulated in a local linear format, and approximated by a neural network (NN). First, the NN is trained offline to provide a good initial estimation of the model, which can directly migrate to the online phase. Then, an adaptive controller is proposed to achieve the shape control tasks, in which the NN is further updated online to compensate for any errors in the offline model caused by insufficient training or changes of DLO properties. The simulation and real-world experiments show that the proposed method can precisely and efficiently accomplish the DLO shape control tasks, and adapt well to new and untrained DLOs.
\end{abstract}

\section{Introduction}

Deformable linear objects (DLOs) refer to deformable objects in one dimension, such as ropes, elastic rods, wires, cables, etc. The demand for manipulating DLOs is reflected in many applications, and a significant amount of research efforts have been devoted to the robotic solutions to these applications. For example, wires are manipulated to assemble devices in 3C manufacturing \cite{8460694}; belts are manipulated in assemblies of belt drive units \cite{jin2021trajectory}; and in surgery, sutures are manipulated to hold tissue together \cite{cao2020sewing}. 

The manipulation tasks of DLOs can be divided into two categories \cite{rita2021reform}. In the first category, the goals are not about the exact shapes of DLOs; rather, they concern high-level conditions such as insertion \cite{weifu2015anonline}, tangling or untangling knots \cite{wakamatsu2006knotting}, obstacle-avoidance \cite{mcconachie2020learning}, flex and flip \cite{jiang2019dynamic},  etc. The second category is about manipulating DLOs to desired shapes, where one key challenge is to obtain the unknown deformation models, i.e., how the robot motion affects the DLO shapes. This paper focuses on the shape control tasks.

Different from rigid objects, it is challenging to obtain the exact models of DLOs, because they are hard to calculate theoretically, and may vary significantly among DLOs. Some analytical physics-based modeling methods can be used to model DLOs, such as mass-spring systems, position-based dynamics, and finite element methods \cite{yin2021modeling}. However, all are approximate models, and require accurate parameters of DLOs which are difficult to acquire. Data-driven approaches have been applied to learn the deformation models, without studying the complex physical dynamics. A common method is to first learn a forward kinematics model offline, and then use model predictive control in manipulation \cite{yan2020learning, wenbo2021deformable,yan_self-supervised_2020,yang2021learning}. Although this allows for learning accurate forward kinematics of an offline-trained DLO, problems may arise when manipulating a different untrained DLO since there is no online update. Reinforcement learning methods have also been studied \cite{lin2020softgym,rita2021learning}, but they are less data-efficient, and the transfer from trained scenarios to untrained scenarios is challenging. Besides these offline approaches, some studies have used purely online methods to estimate the local linear deformation model of manipulated DLOs, which can be applied to any new DLO \cite{navarro2016Automatic,zhu_dual-arm_2018, jin2019robust, lagneau_automatic_2020}. However, these online estimated models are less accurate because only a small amount of local data can be utilized.

\begin{figure}[tb]
    \centering
    \includegraphics[width=8.5cm]{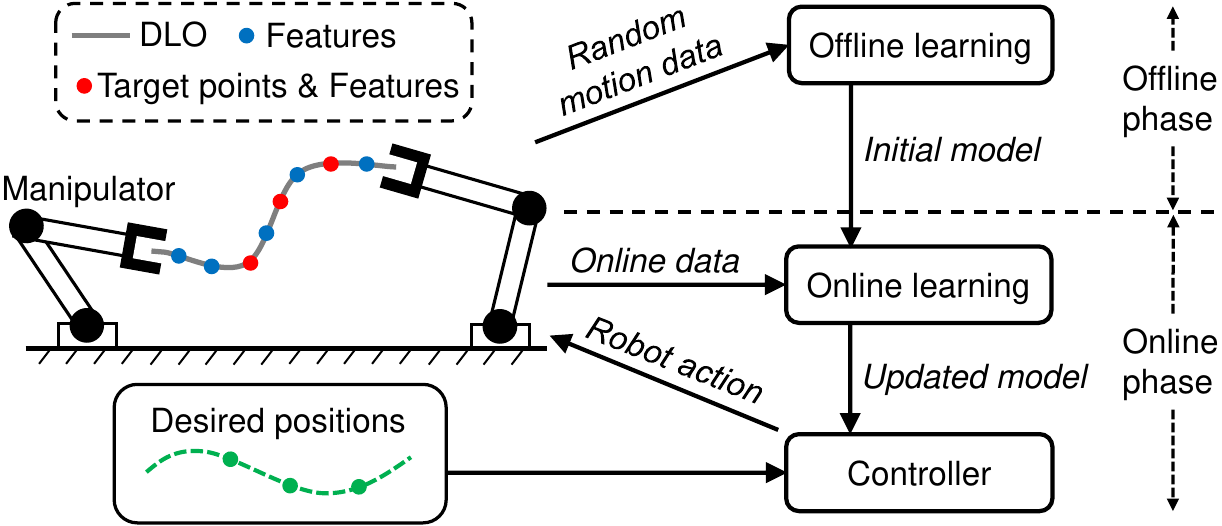}
    \vspace{-2mm}
    \caption{Overview of the proposed scheme for DLO shape control. The shape of the DLO is represented by multiple features along the DLO. Some of the features are chosen as target points, and the task is defined as moving the target points to their desired positions. In the offline phase, an initial estimation of the deformation model is learned. Then, in the online phase, the shape control task is executed, and the model is further updated to compensate for offline modeling errors.}
    \label{fig:overview}
    \vspace{-7mm}
\end{figure}


In this paper, we propose a scheme for the shape control of DLOs, where the unknown deformation model is estimated with both offline and online learning, shown in Fig. \ref{fig:overview}. It allows more accurate modeling through offline learning and further updating for specific DLOs via online learning during manipulation. Specifically, we use a radial-basis-function neural network (RBFN) to model the mapping from the current state to the current local linear deformation model. In the offline phase, the RBFN is trained on the collected data. The offline model then directly migrates to the online phase as an initial estimation. In the online manipulation phase, an adaptive controller is proposed to control the shape, in which the RBFN is further updated to adapt to the manipulated DLO concurrently. Thus, the offline learning and online learning complement each other. In addition, we apply proper state representations and domain randomization methods to improve the model's generalization ability. The simple structure of RBFN and the linear format of the deformation model enable data-efficient offline training and fast online adaptation. Compared to the nonlinear offline methods, our method is more data-efficient and can adapt to untrained DLOs; compared to the purely online methods, our method is more stable and can handle more complex tasks. The stability of the closed-loop system and the convergence of task errors are analyzed using the Lyapunov method. Simulation and real-world experiment results are presented to demonstrate the better performance of the proposed scheme compared to the previous data-driven methods. The video and code are available at \url{https://mingrui-yu.github.io/shape_control_DLO/}.

\section{Related Work}

In this section, existing approaches for the shape control of DLOs will be discussed. 

Analytical physics-based modeling of DLOs has been researched over the past several decades \cite{yin2021modeling}. Some works about shape control were based on physics-based models. In \cite{duenser2018interactive}, finite element model (FEM) simulation of DLOs was used for open-loop shape control. An approach using reduced FEM to closed-loop shape control of DLOs was proposed in \cite{koessler2021anefficient}. These methods highly rely on the accuracy of the analytical model, requiring the accurate DLO parameters which are hard to obtain in reality. 

Data-driven approaches have been applied to the shape control of DLOs recently, dispensing with analytical modeling. In \cite{rambow2012autonomous,nair2017combining,tang2018aframework}, the shaping of DLOs was addressed by learning from human demonstrations. Robots could reproduce human actions for specific tasks. 
Reinforcement learning (RL) has also been applied to learn policies for DLO shape control in an end-to-end manner. 
A simulated benchmark of RL algorithms for deformable object manipulation was presented in \cite{lin2020softgym}, in which the Soft Actor Critic (SAC) algorithm performs best in rope straightening tasks. RL policies for shape control of elastoplastic DLOs were learned in \cite{rita2021learning}.
Like other RL applications, these methods suffer from high training expenses and challenging transfer from simulation to real-world scenarios.

Different from the end-to-end methods, many works first learn forward kinematics models of DLOs offline, and then use model predictive control (MPC) to control the shape. The forward kinematics model predicts the shape at the next time step based on the current shape and input action. In \cite{yan2020learning, wenbo2021deformable}, an encoder from the image space to the latent space, and a forward kinematics model in the latent space, were jointly trained. A more robust and data-efficient approach is to estimate the DLO state first and then learn the forward kinematics in the physical state space. A bi-directional LSTM network whose structure is similar to chain-like DLOs was applied in \cite{yan_self-supervised_2020}. An interaction network was integrated with a bi-directional LSTM network in \cite{yang2021learning} to better learn the local interactions between segments of DLOs. The problem with these offline methods is that the generalization to different untrained DLOs cannot be guaranteed.

To control the shape of unknown objects, a series of methods tackle the shape control problem based on purely online estimation of the local linear deformation models of DLOs, in which a small change of the DLO is linearly related to a small displacement of the robot by a locally effective estimated Jacobian matrix. The control input can be directly calculated using the inverse of the Jacobian matrix. 
In \cite{zhu_dual-arm_2018, jin2019robust, lagneau_automatic_2020}, the local Jacobian matrix was obtained using the (weighted) least square estimation on only the data in the current sliding window. However, the accuracy of the online estimated models is limited and cannot be improved with more data. Thus, these methods mostly handle tasks with local and small deformation.



\section{Methodology}

This paper considers the quasi-static shape control of elastic DLOs. `Quasi-static' refers to the motion being slow, in which the shapes of DLOs are assumed to be determined by only their potential energies and no inertial effects \cite{navarro2016Automatic}. As illustrated in Fig. \ref{fig:overview}, the robot end-effectors grasp the ends of the DLO and manipulate it to the desired shape. The overall shape of the DLO is represented by the positions of multiple features uniformly distributed along the DLO. The target points are chosen from the features, and the task is defined as moving the target points on the DLO to their corresponding desired positions. The specific choice of the target points depends on the task needs.

Some frequently-used notations are listed as follows. The vertical concatenation of column vector $\bm a$ and $\bm b$ is denoted as $[\bm a; \bm b]$. The position vector of the end-effectors is represented as $\bm r \in \Re^{n}$. The position of the $i^{\rm th}$ feature is represented as $\bm x_i \in \Re^l$. The overall shape vector of the DLO is represented as $\bm{x} = [\bm{x}_1; \cdots; \bm{x}_m] \in \Re^{lm}$ , where $m$ is the number of the features. The dimension $n$ and $l$ are adjustable according to the requirements of the task.

\subsection{Local Linear Deformation Model}
One key problem of the shape control of DLOs is to study the mapping from the motion of the end-effectors to the motion of the DLO features. The velocity vector of the DLO features can be locally linearly related to the velocity vector of the end-effectors using a Jacobian matrix \cite{zhu_dual-arm_2018,lagneau_automatic_2020, jin2019robust, navarro2016Automatic}. Different from the previous works, we estimate the Jacobian matrix by learning the mapping from the current state $(\bm x, \bm r)$ to the current Jacobian matrix $\bm J$, i.e., $\bm J$ can be obtained as a function of $\bm x$ and $\bm r$: 
\begin{equation} \label{Jacob1}
    \dot{\bm x} = \bm J(\bm x, \bm r) \dot{\bm r} 
\end{equation}

\textit{Proposition 1:} With the quasi-static assumption, the velocity vector of the features on the elastic DLO can be related to the velocity vector of the end-effectors as (\ref{Jacob1}).


\textit{Proof:} Denote the potential energy of the elastic DLO as $E$, which is assumed to be fully determined by $\bm x$ and $\bm r$. 
In the quasi-static assumption, internal equilibrium holds at all states during the manipulation, where the DLO's internal shape $\bm x$ locally minimizes the potential energy $E$ \cite{bretl2014quasi}. That is, $\partial E / \partial \bm x \hspace{-0.5mm}=\hspace{-0.5mm} \bm 0$ at any state.
Consider the DLO is moved from  state $(\bar{\bm x}, \bar{\bm r})$ to  state $(\bar{\bm x} + \delta \bm x, \bar{\bm r} + \delta \bm r)$ where $\delta \bm x$ and $\delta \bm r$ are small displacements of the features and the end-effectors. Denote $\partial E / \partial \bm x$ as $\bm g(\bm x, \bm r)$, $\partial^2 E / (\partial\bm x \partial\bm x)$ as $\bm A(\bm x, \bm r)$, and $\partial^2 E / (\partial\bm x \partial\bm r)$ as $\bm B(\bm x, \bm r)$. Using Taylor expansion and neglecting higher order terms, we have
\begin{equation}
    \bm g(\bar{\bm x} \hspace{-0.3mm}+\hspace{-0.3mm} \delta \bm x, \bar{\bm r} \hspace{-0.3mm}+\hspace{-0.3mm} \delta \bm r) \approx \bm g(\bar{\bm x}, \bar{\bm r}) + \bm A(\bar{\bm x}, \bar{\bm r}) \delta {\bm x} + \bm B(\bar{\bm x}, \bar{\bm r}) \delta {\bm r}
\end{equation}
where $\bm g(\bar{\bm x} \hspace{-0.3mm}+\hspace{-0.3mm} \delta \bm x, \bar{\bm r} \hspace{-0.3mm}+\hspace{-0.3mm} \delta \bm r) = \bm g(\bar{\bm x}, \bar{\bm r}) = \bm 0$. Note that $\bm A$ and $\bm B$ physically represent the unknown stiffness matrices. Assuming the DLO has a positive and full-rank stiffness matrix around the equilibrium point, matrix $\bm A$ is invertible \cite{navarro2016Automatic}. Then, it can be obtained that
\begin{equation}
    \delta {\bm x} \approx -  \left(\bm A(\bar{\bm x}, \bar{\bm r})\right)^{-1} \bm B(\bar{\bm x}, \bar{\bm r}) \delta {\bm r}
\end{equation}
In slow manipulations, $\dot{\bm x} \approx \delta \bm x / \delta t$ and $\dot{\bm r} \approx \delta \bm r / \delta t$ with small $\delta t$. Then, denoting $-(\bm A(\bm x, \bm r))^{-1} \bm B(\bm x, \bm r)$ as $\bm J(\bm x, \bm r)$, we derive (\ref{Jacob1}) and prove the proposition.

Note that (\ref{Jacob1}) can be rewritten as
\begin{equation} \label{Jacob2}
    \dot{\bm x} = \left[ \begin{array}{c}
      \dot{\bm{x}}_1   \\ \vdots \\ \dot{\bm{x}}_m \end{array} \right]
    = \left[ \begin{array}{c}
      \bm{J}_1(\bm x, \bm r)   \\ \vdots \\ \bm{J}_m(\bm x, \bm r) \end{array} \right]
      \dot{\bm r}
\end{equation}
where $\bm{J}_k(\bm x, \bm r)$ is the $((k-1)l+1)^{\rm th}$ to $(kl)^{\rm th}$ rows of $\bm J(\bm x, \bm r)$. Thus, it can be obtained that
\begin{equation} \label{Jacob3}
    \dot{\bm x}_k = \bm J_k(\bm x, \bm r) \dot{\bm r}, \quad k =  1,\cdots , m
\end{equation}
which indicates that different features correspond to different Jacobian matrices. This formulation makes it convenient when choosing any subset of features as the target points in the manipulation task.

However, it is difficult to theoretically calculate the Jacobian matrix. We estimate the Jacobian matrix in a data-driven way, combining both offline learning and online learning.

\subsection{Offline Learning}

Prior to the shape control tasks, a data-driven learning method is employed to obtain the initial estimation of the model, based on offline collected data.

\begin{figure}[tb]
    \centering
    \includegraphics[width=8.5cm]{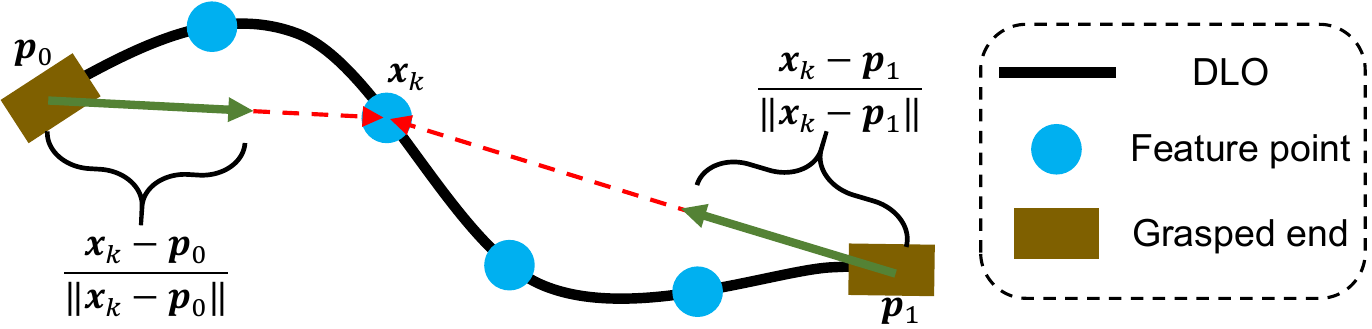}
    \vspace{-3mm}
    \caption{An illustration of the proposed DLO state representation in (\ref{equation:DLO_state_representation}). The position of the $k^{\rm th}$ feature $\bm x_k$ can be determined by $\bar{\bm x}_k = \left[ (\bm x_k - \bm p_0)/\| \bm x_k - \bm p_0 \|; (\bm x_k - \bm p_1)/\| \bm x_k - \bm p_1 \|  \right]$, if the positions of the left and right ends ($\bm p_0$ and $\bm p_1$) are given. In (\ref{equation:DLO_state_representation}), the positions of the ends are described by $\bar{\bm p} = (\bm p_1 - \bm p_0)/\| \bm p_1 - \bm p_0 \|$, so the overall translation and scale are ignored in the representation.}
    \label{fig:state_representation}
    \vspace{-5mm}
\end{figure}

We apply a neural network (NN) to approximate the Jacobian matrix, in which the input is the current state and the output is the Jacobian. Two properties of the Jacobian can be noticed intuitively: (1)translation-invariance: translation of the whole DLO without changes of the shape will not alter the Jacobian matrix; (2)approximate scale-invariance: DLOs with different lengths but similar overall shapes and the same number of features may have similar Jacobian matrices. Thus, to improve the NN's generalization ability, we modify the representation of the input state from $[\bm x; \bm r]$ to
\begin{equation} \label{equation:DLO_state_representation}
    \bm \phi \stackrel{\triangle}{=} \left[ \bar{\bm{x}}_1; \cdots; \bar{\bm{x}}_m; \hspace{1mm} \bar{\bm p}; \bm q_0; \bm q_1 \right]
\end{equation}
where 
\begin{equation}
    \bar{\bm{x}}_k = \left[ \frac{\bm x_k - \bm p_0}{\| \bm x_k - \bm p_0 \|}; \frac{\bm x_k - \bm p_1}{\| \bm x_k - \bm p_1 \|}  \right], \quad \bar{\bm p} = \frac{\bm p_1 - \bm p_0}{\| \bm p_1 - \bm p_0 \|}
\end{equation}
$(k = 1, \cdots, m)$, where $\bm p_0$, $\bm p_1$ are the positions of the left and right grasped ends of the DLO, and $\bm q_0$, $\bm q_1$ are the orientations of the left and right grasped ends. As illustrated in Fig. \ref{fig:state_representation}, this relative representation only determines the overall shape, ignoring the scale and overall translation. Therefore, it is much more data-efficient than the absolute representation $[\bm x; \bm r]$ which requires a larger network and training dataset to guarantee the generalization to different DLOs. Note that this representation avoids using relative positions between adjacent features, which may suffer from accumulated errors when perception errors exist.

Then, (\ref{Jacob3}) can be rewritten as
\begin{equation} \label{Jacob4}
    \dot{\bm x}_k = \bm J_k(\bm \phi) \dot{\bm r}, \quad k =  1,\cdots , m
\end{equation}

We apply a radial-basis-function neural network (RBFN) to represent the actual Jacobian matrix as a function of $\bm \phi$:
\begin{equation} \label{vecJ=W*theta}
    {\rm vec} \left(\bm J_k(\bm \phi)\right) = \bm W_k \bm{\theta}(\bm \phi), \quad k =  1,\cdots , m
\end{equation}
where ${\rm vec}(\cdot)$ refers to the column vectorization operator, and $\bm W_k$ is the matrix of unknown actual weights of the RBFN for the $k^{\rm th}$ feature. The $\bm \theta (\bm \phi)$ represents the vector of activation functions, and $\bm{\theta}(\bm{\bm \phi})=[\theta_1(\bm{\bm \phi}), \theta_2(\bm{\bm \phi}), \cdots, \theta_q(\bm{\bm \phi})]^T \in \Re^q $. We use gaussian radial function as the activation function: 
\begin{equation} \label{singleNeuron}
\theta_i(\bm{\phi})={\rm e}^{\frac{-||\bm{\phi} -\bm \mu_i||^2}{\sigma_i^2}}, \quad i =  1,\cdots , q
\end{equation}
where the parameters $\bm \mu_i$ and $\sigma_i$ are trainable in the offline phase but fixed in the online phase.

Equation (\ref{vecJ=W*theta}) can be decomposed as
\begin{equation} \label{equation:J=Wtheta}
    \bm{J}_{ki}(\bm{\phi}) = \bm{W}_{ki} \bm{\theta}(\bm{\phi}), \quad i =  1,\cdots , n
\end{equation}
where $\bm{J}_{ki}$ is the $i^{\rm th}$ column of $\bm J_k$, and $\bm{W}_{ki}$ is the ${((i-1) l+1)}^{\rm th}$ to ${(i l)}^{\rm th}$ rows of $\bm{W}_k$. Subscribing (\ref{equation:J=Wtheta}) into (\ref{Jacob4}) yields
\begin{equation}
    \dot{\bm{x}}_k = \bm{J}_k(\bm{\phi})\dot{\bm{r}} = \sum_{i=1}^{n}\bm{J}_{ki}(\bm{\phi})\dot{r}_i = \sum_{i=1}^{n}\bm{W}_{ki}\bm{\theta}(\bm{\phi})\dot{r}_i
\end{equation}
where $\dot{r}_i$ is the $i^{\rm th}$ element of $\dot{\bm r}$.

The estimated Jacobian matrix is represented as 
\begin{equation} \label{estimatedJ}
    {\rm{vec}} (\hat{\bm{J}}_k(\bm{\phi})) = \hat{\bm{W}}_k\bm{\theta}(\bm{\phi})
\end{equation}
where $\hat{\bm{W}}$ is the matrix of estimated weights. The approximation error for the $k^{\rm th}$ feature $\bm{e}_k$ is specified as 
\begin{equation} \label{ew}
\begin{aligned}
    \bm{e}_k &= \dot{\bm{x}}_k - \hat{\bm{J}}_k(\bm{\phi})\dot{\bm{r}}
    \\
    &= \sum_{i=1}^{n}\bm{W}_{ki}\bm{\theta}(\bm{\phi})\dot{r}_i - \sum_{i=1}^{n}\hat{\bm{W}}_{ki}\bm{\theta}(\bm{\phi})\dot{r}_i = \sum_{i=1}^{n}\Delta{\bm{W}}_{ki}\bm{\theta}(\bm{\phi})\dot{r}_i
\end{aligned}
\end{equation}

\begin{figure}[tb]
    \centering
    \includegraphics[width=6.5cm]{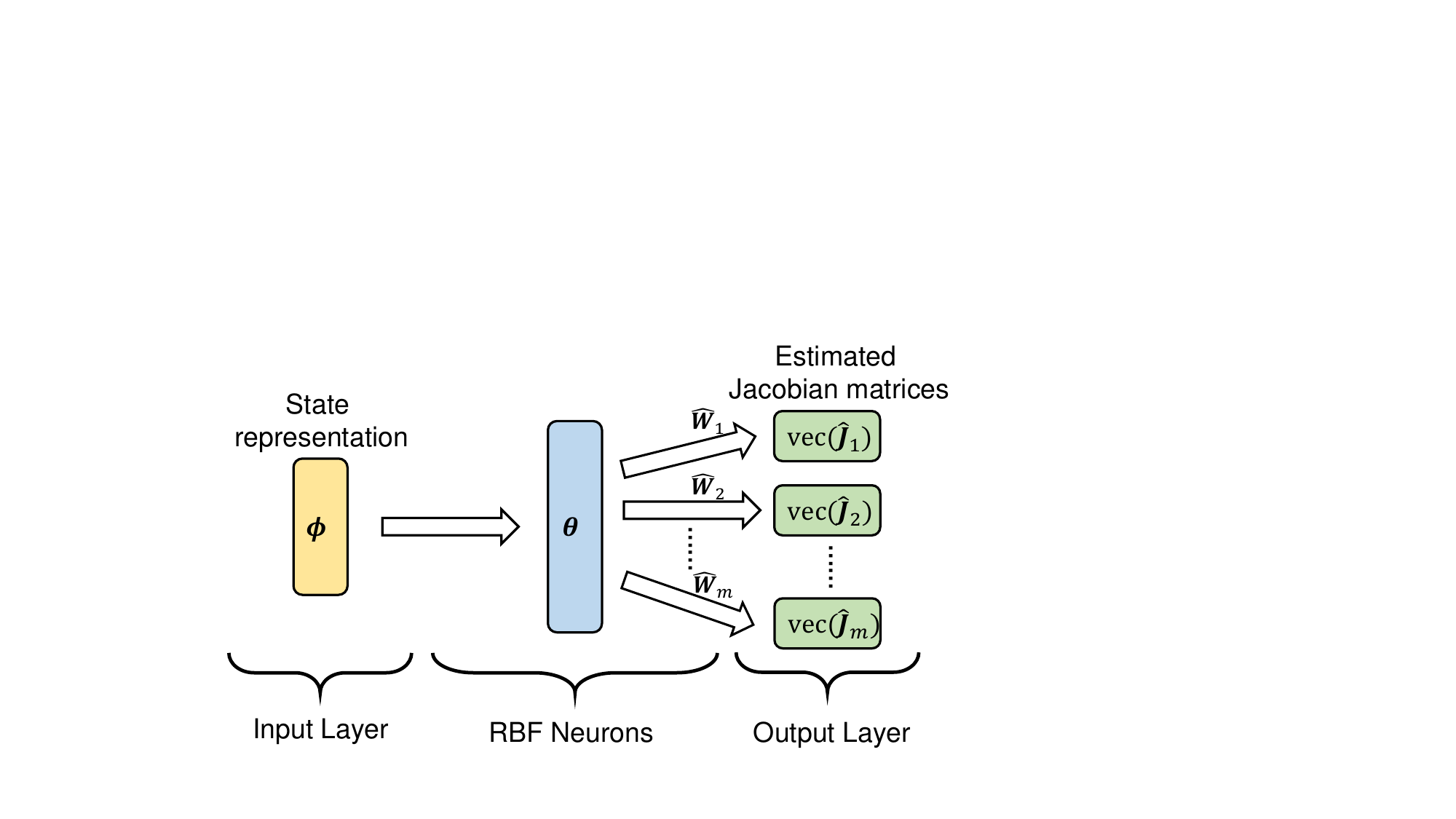}
    \vspace{-3mm}
    \caption{The architecture of the RBFN for learning the local linear deformation model. The network takes the state representation in (\ref{equation:DLO_state_representation}) as the input and outputs the estimated Jacobian matrices which relate the velocity vectors of DLO features to the velocity vector of the robot end-effectors.}
    \label{fig:RBFN}
    \vspace{-5mm}
\end{figure}

The architecture of the RBFN is shown in Fig. \ref{fig:RBFN}. Note that the learning or estimation for Jacobian matrices of different features is carried out in parallel. In the offline phase, the ends of the DLO are controlled to move randomly to collect the training dataset, which contains $\bm x_k, \dot{\bm{x}}_k, \bm r, \dot{\bm{r}}, (k=1,\cdots,m)$. Then, the RBFN is trained on the collected data.  Considering the noise and outliers in the data, we use the smooth $L1$ loss \cite{girshick2015fast} of $\bm e_k$ for training.

The k-means clustering on a subset of the training data is used to calculate the initial value of $\bm{\mu}_i$ and $\sigma_i, (i = 1, \cdots,q)$. Then, all parameters including $\bm{\mu}_i$, $\sigma_i$ and $\hat{\bm{W}}$ are updated using the {\em Adam} optimizer \cite{kingma2014adam}. We choose RBFN for its simple structure, robustness, and online learning ability \cite{yu2011advantages}. Though less expressive than some more complex network architectures, it performs well enough in this work.


\subsection{Adaptive Control through Online Learning}
Considering the differences between the manipulated DLO in the online phase and the trained DLOs in the offline phase, online learning during manipulation is required. We propose an adaptive control scheme, in which the offline estimated model is treated as an initial approximation and then further updated during the shape control tasks.

The control objective is to move the target points on the DLO to the desired positions. The target points can be any subset of the features, whose indexes form set $\mathcal{C}$. Then, the target shape vector $\bm x^c$, target Jacobian matrix $\bm J^c(\bm{\phi})$, and target weights $\bm W^c$ are denoted as
\begin{equation} \label{equation:all_c}
    \bm{x}^c \stackrel{\triangle}{=} \left[\hspace{-1mm} \begin{array}{c}
      \vdots   \\ \bm x_k \\ \vdots \end{array} \hspace{-1mm}\right] \hspace{-1mm}, 
    \bm{J}^c(\bm{\phi}) \stackrel{\triangle}{=}  \left[\hspace{-1mm} \begin{array}{c}
      \vdots   \\ \bm J_k(\bm{\phi}) \\ \vdots \end{array} \hspace{-1mm}\right]
      \hspace{-1mm}, 
    \bm{W}^c \stackrel{\triangle}{=}  \left[\hspace{-1mm} \begin{array}{c}
      \vdots   \\ \bm W_k \\ \vdots \end{array} \hspace{-1mm}\right]
  \hspace{-1mm},  
  k \in \mathcal{C}
\end{equation}

The velocities of the robot end-effectors $\dot{\bm r}$ are controlled, and the control input is specified as
\begin{equation} \label{control_law}
\dot{\bm r} = - \alpha \left(\hat{\bm J}^c(\bm{\phi})\right)^{\dagger} \Delta\bm x^c
\end{equation}
where $\left(\hat{\bm J}^c(\bm{\phi})\right)^{\dagger}$ is the Moore-Penrose pseudo-inverse of the estimated Jacobian matrix. In addition, $\Delta \bm x^c = \bm x^c - \bm x^c_{desired}$ where $\bm x^c_{desired}$ is the desired position vector of the target points, and $\alpha \in \Re$ is a positive control gain. In actual implementations, $\dot{\bm r}$ is bounded to avoid too fast motion.

The online updating law of the $j^{\rm th}$ row of $\hat{\bm W}_{ki}$ of the RBFN is specified as 
\begin{equation} \label{control_update}
\dot{\hat{\bm W}}_{kij}^T = \hspace{0.1cm} \dot{r}_i\bm\theta(\bm{\phi}) (\eta_1 \Delta x_{kj} + \eta_2 e_{kj}), \quad j=1 ,\cdots, l
\end{equation}
where $\Delta x_{kj}$ is the $j^{\rm th}$ element of the task error $\Delta \bm{x}_k$, and $e_{kj}$ is the $j^{\rm th}$ element of the approximation error $\bm e_k$. The $\eta_1$ and $\eta_2$ are positive scalars. Such updating is done for all $k \in \mathcal{C}$ and $i = 1, \cdots, n$. 

The proposed control scheme by (\ref{control_law}) and (\ref{control_update}) allows controlling the target points on the DLO to the desired positions while updating the RBFN concurrently to compensate for any offline modeling errors. 

The stability of the system is analyzed as follows. Below ${\bm J}^c(\bm{\phi}), \bm J_k(\bm{\phi}), {\bm \theta}(\bm{\phi})$ are shortened to ${\bm J}^c, {\bm J}_k, \bm \theta$ for simplicity. Premultiplying both sides of (\ref{control_law}) by $\hat{\bm J}^c$, we have
\begin{equation} \label{control_loop_equation_2}
\hat{\bm J}^c\dot{\bm r} = - \alpha \hat{\bm J}^c (\hat{\bm J}^c)^{\dagger}\Delta\bm x^c
\end{equation}
Note that from (\ref{ew}) and (\ref{equation:all_c}), it can be obtained that 
\begin{equation} \label{control_loop_equation_3}
    \hat{\bm J}^c \dot{\bm{r}} 
    = \hat{\bm J}^c \dot{\bm{r}} - {\bm J}^c \dot{\bm{r}} +{\bm J}^c \dot{\bm{r}} 
    = - \bm e^c + \dot{\bm x}^c
\end{equation}
where $\bm e^c = [\cdots; \bm e_k; \cdots], k \in \mathcal{C}$. Since the desired positions are fixed, substituting (\ref{control_loop_equation_3}) into (\ref{control_loop_equation_2}) yields
\begin{equation} \label{control_loop_equation_4}
\bm e^c = \Delta\dot{\bm x}^c + \alpha \hat{\bm J}^c (\hat{\bm J}^c)^{\dagger} \Delta\bm x^c
\end{equation}
A Lyapunov-like candidate is given as 
\begin{equation} \label{equation:V}
    V = \frac{1}{2} (\Delta\bm x^c)^T \Delta\bm x^c + \frac{1}{2\eta_1} \sum_{k\in \mathcal{C}}\sum\limits_{i=1}^n\sum\limits_{j=1}^l \Delta\bm W_{kij} \Delta\bm W_{kij}^T
\end{equation}
Differentiating (\ref{equation:V}) with respect to time and substituting (\ref{control_loop_equation_4}) (\ref{control_update}) and (\ref{ew}) into it, we can obtain that
\begin{equation} \label{equation:dotV}
\begin{aligned}
    & \dot V 
    = (\Delta \bm x^c)^T \Delta \dot{\bm x}^c - \sum_{k\in \mathcal{C}}\sum\limits_{i=1}^n\sum\limits_{j=1}^l \frac{1}{\eta_1} \Delta\bm W_{kij}  \dot{\hat{\bm W}}_{kij}^T \\
    &= - \alpha (\Delta \bm x^c)^T \hat{\bm J}^c (\hat{\bm J}^c)^{\dagger} \Delta\bm x^c + (\Delta \bm x^c)^T \bm e^c \\
    &\quad - \sum_{k\in \mathcal{C}}\sum\limits_{i=1}^n\sum\limits_{j=1}^l \frac{1}{\eta_1} \Delta\bm W_{kij} \left[\dot{r}_i\bm\theta (\eta_1 \Delta x_{kj} + \eta_2 e_{kj})\right] \\
    &= - \alpha (\Delta \bm x^c)^T \hat{\bm J}^c (\hat{\bm J}^c)^{\dagger} \Delta\bm x^c + (\Delta \bm x^c)^T \bm e^c \\ 
    & \quad - (\bm e^c)^T \Delta \bm x^c - \frac{\eta_2}{\eta_1} (\bm e^c)^T \bm e^c \\
    &= - \alpha (\Delta \bm x^c)^T \hat{\bm J}^c (\hat{\bm J}^c)^{\dagger} \Delta\bm x^c - \frac{\eta_2}{\eta_1} (\bm e^c)^T \bm e^c 
    \leq 0
\end{aligned}
\end{equation}
As $V \hspace{-0.5mm} > \hspace{-0.5mm} 0$ and $\dot V \hspace{-0.5mm} \leq \hspace{-0.5mm} 0$, the closed-loop system is stable. The boundedness of $V$ ensures the boundedness of $\Delta \bm x^c$ from (\ref{equation:V}). If $l  \hspace{-0.5mm} \times \hspace{-0.5mm} |\mathcal{C}| \hspace{-0.5mm} \leq \hspace{-0.5mm} n$ and $\hat{\bm J}^c$ holds full row rank, $\hat{\bm J}^c (\hat{\bm J}^c)^{\dagger}$ is an identity matrix. Then, it can be proved that $\Delta \bm x^c \hspace{-0.5mm} \rightarrow \hspace{-0.5mm} \bm 0$ as $t \hspace{-0.5mm} \rightarrow \hspace{-0.5mm} \infty$, following \cite{arimoto1996control}. Otherwise, $\hat{\bm J}^c (\hat{\bm J}^c)^{\dagger}$ is only positive semi-definite, resulting in an underactuation system. However, from (\ref{ew}), (\ref{control_law}) and (\ref{equation:dotV}) it can be proved that $\dot V \hspace{-0.5mm} = \hspace{-0.5mm} 0$ if and only if $\dot{\bm r} \hspace{-0.5mm} = \hspace{-0.5mm} \bm 0$, which only happens when there are huge conflicts between the desired moving directions of different target points so that the robot movement in any direction cannot reduce the $\Delta \bm x^c$ at this "local minimum point". Actually, this situation happens rarely in experiments owing to the coupling between the target points, so in most cases $\dot{V} \hspace{-0.5mm} < \hspace{-0.5mm} 0$ always holds and finally $\Delta \bm x^c \hspace{-0.5mm} \rightarrow \hspace{-0.5mm} \bm 0$.

\section{Results}

\begin{figure} [tb]
  \centering 
  \subfigure[2D]{ 
    \includegraphics[width=3.6cm]{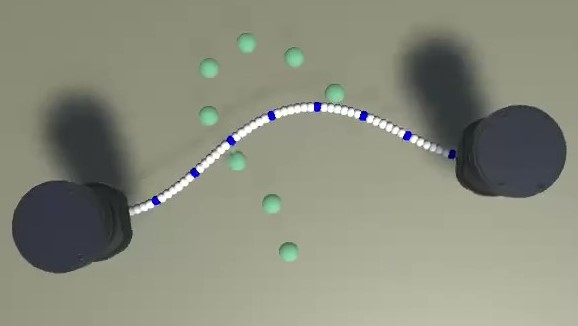} 
  } 
  \subfigure[3D]{ 
    \includegraphics[width=3.6cm]{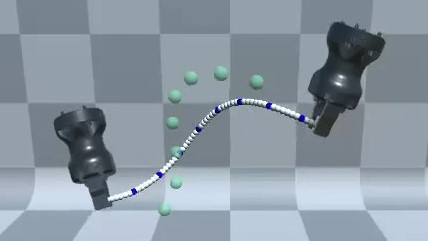} 
  }
  \vspace{-2mm}
  \caption{The simulation environment for the DLO shape control tasks. The blue points represent the features along the DLO. The green points represent the desired positions of the features.}
  \label{fig:simulation_environtment}
    \vspace{-5mm}
\end{figure}

We carry out both simulation and real-world experiments to validate the proposed method. The simulation of DLOs is based on Obi \cite{obi}, a unified particle physics engine for deformable objects in Unity3D \cite{unity}, as shown in Fig. \ref{fig:simulation_environtment}. In the simulation, the two ends of the DLO are grasped by two grippers, which can translate and rotate. Both 2D and 3D tasks are tested. In the 2D tasks, the environment dimension $l$ is $2$ and the control input dimension $n$ is $6$; in the 3D tasks $l\hspace{-1mm}=\hspace{-1mm}3, n\hspace{-1mm}=\hspace{-1mm}12$. In the real-world experiments, the DLOs are placed on a table. One end of the DLO is grasped by a UR5 arm, and the other is fixed. Thus, $l\hspace{-1mm}=\hspace{-1mm}2$ and $n\hspace{-1mm}=\hspace{-1mm}3$. The shape of the DLO is represented by 8 features ($m\hspace{-1mm}=\hspace{-1mm}8$), and the positions of the features are obtained by measuring the markers on the DLO with a calibrated RGB camera in the experiments. Both the data collection frequency and control frequency are 10 Hz.

We choose three representative classes of methods for comparison. The first class is learning forward kinematics models of DLOs offline and using MPC for shape control (FKM+MPC). According to \cite{yan_self-supervised_2020}, we choose bi-directional LSTM (biLSTM) for modeling and Model Predictive Path Integral Control (MPPI) for control. The second class is estimating the Jacobian matrix online using weighted least square estimation (WLS). We specifically use the method in \cite{lagneau_automatic_2020}. The third is based on reinforcement learning. We train an agent using Soft Actor Critic (SAC) \cite{haarnoja2018soft}.

\subsection{Offline Learning of the Deformation Model}

The offline data of DLOs are collected in simulation, by randomly moving the ends of the DLOs. A RBFN with 256 neurons in the middle layer ($q=256$) is first trained offline to learn the initial deformation model.

\begin{figure} [tb]
  \centering 
  \subfigure[2D]{ 
    \includegraphics[width=4.1cm]{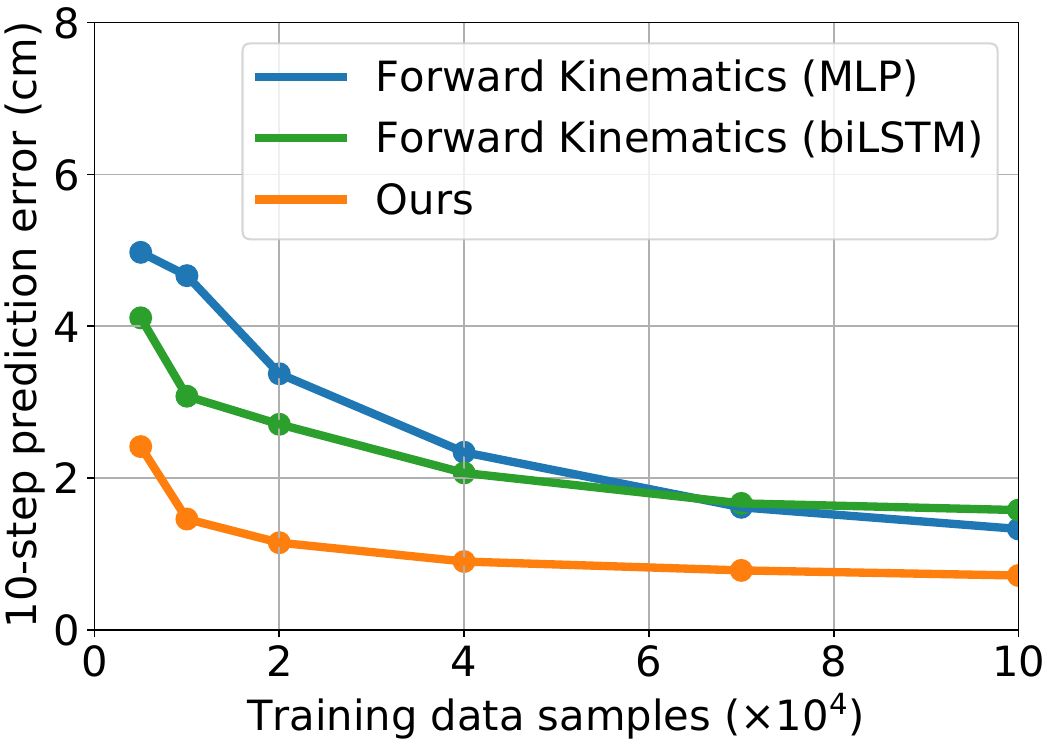} 
  } 
  \hspace{-0.4cm}
  \subfigure[3D]{ 
    \includegraphics[width=4.1cm]{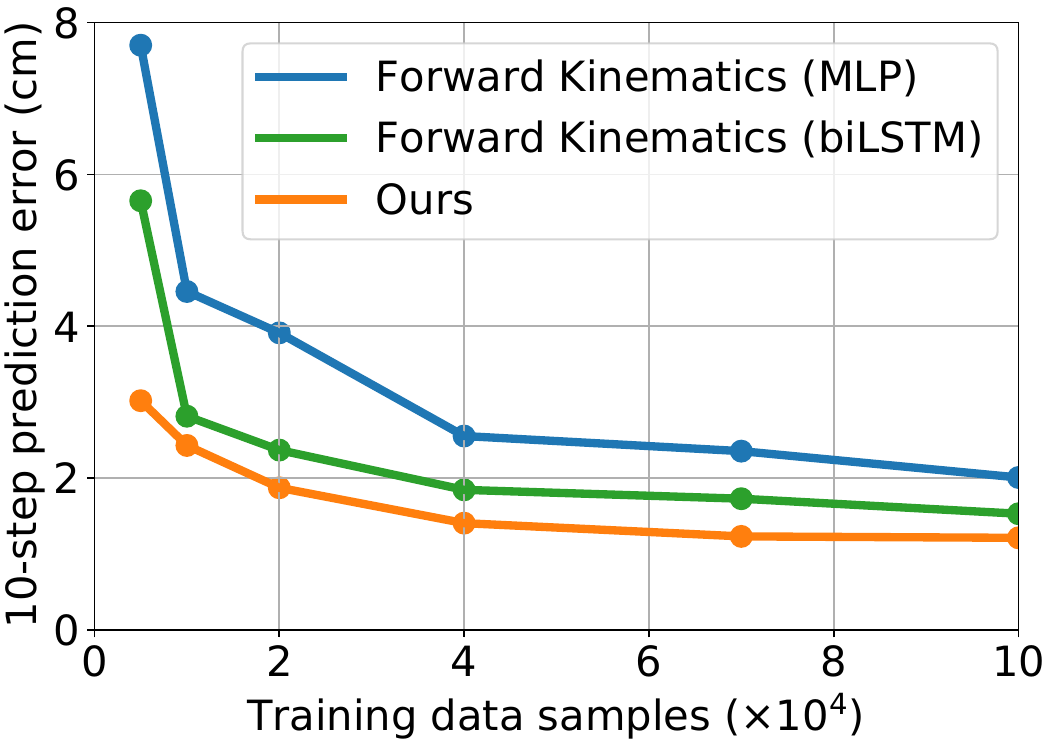} 
  }
  \vspace{-3mm}
  \caption{The relationship between the offline modeling accuracy and the amount of training data. Training data samples and 10k test samples are collected on the same DLO. The error is the average Euclidean distance between the prediction and ground truth of the shape after 10 steps.}
  \label{fig:exp_offline}
    \vspace{-2mm}
\end{figure}

First, we test the offline modeling accuracy on a certain DLO and its relationship to the amount of training data, in which we compare our local linear Jacobian model with nonlinear forward kinematics models based on multi-layer perceptrons (MLP) or biLSTM. Training data and 10k test data are collected on the same DLO. For testing, we use trained models to predict the shape of the DLO after 10 steps. The prediction of the shape at the next step using our method is calculated as $\hat{\bm x}_{[t+1]} = \bm x_{[t]} + \Delta t \hspace{1mm} \hat{\bm J}(\bm \phi_{[t]}) \dot{\bm r}_{[t]}$, where the subscript $[t]$ represents the variables at step $t$ and $\Delta t$ is the step interval. Shown in Fig. \ref{fig:exp_offline}, the results indicate that our Jacobian model can achieve higher prediction accuracy with less training data. The biLSTM-based model incorporates the physics priors of chain-like DLOs \cite{yan_self-supervised_2020}, so it performs better than MLP when the training set is small. Our Jacobian model implies strong local linear prior, which is theoretically and practically reasonable. Hence, the learning efficiency is highly improved. 

Second, we further compare the performance of our methods using the absolute state input $[\bm x; \bm r]$ and relative state input $\phi$ as (\ref{equation:DLO_state_representation}) in the RBFN. We collect data of 10 different DLOs in the simulation. For testing, we perform cross validation, i.e., for each round the model is trained on $9\hspace{-0.1mm}\times\hspace{-0.1mm}3$k data of 9 DLOs and tested on 10k data of the remaining one. We also test the performance on the test set with constant position translation. The average results of 10 rounds shown in Table \ref{tab:exp_abs_state_rel_state} reveal that the relative state input can achieve higher prediction accuracy on new DLOs with different lengths. In addition, when position translation is added, the relative state input is not affected, while the performance of the absolute state input significantly decreases.

\begin{table}[tb]
\centering
\caption{Comparison of the absolute state input and relative state input in our method.}
\vspace{-2mm}
\label{tab:exp_abs_state_rel_state}
\begin{threeparttable}[b]
\begin{tabular}{@{}c|cc|cc@{}}
\toprule
\multirow{2}{*}{\begin{tabular}[c]{@{}c@{}}Feature velocity\\ prediction\\ relative error\tnote{a}\end{tabular}} & \multicolumn{2}{c|}{2D}                                                                 & \multicolumn{2}{c}{3D}                                                                  \\
                                                                                                          & New DLO\tnote{b}$\hspace{-3mm}$ & \begin{tabular}[c]{@{}c@{}}New DLO\\ +translation\tnote{c}\end{tabular} & New DLO$\hspace{-3mm}$ & \begin{tabular}[c]{@{}c@{}}New DLO\\ +translation\end{tabular} \\ \midrule
Absolute state input                                                                                      & 30.0\%                 & 54.5\%                                                         & 41.1\%                 & 81.4\%                                                         \\
Relative state input                                                                                      & \textbf{25.9\%}        & \textbf{25.9\%}                                                & \textbf{36.2\%}        & \textbf{36.2\%}                                                \\ \bottomrule
\end{tabular}
\begin{tablenotes}
     \item[a] The relative error is calculated as the average of $\|\dot{\bm x} - \hat{\bm J} \dot{\bm r}\| / \|\dot{\bm x}\|$. 
    \item[b] A new DLO whose length may be different from the trained DLOs'. 
    \item[c] A 0.2m position translation offset is added in each dimension. 
\end{tablenotes}
\end{threeparttable}
\vspace{-5mm}
\end{table}

\begin{table*}[tb]
\centering
\caption{Performance of the methods in 2D and 3D DLO shape control tasks in simulation.}
\vspace{-2mm}
\label{tab:sim_control_results}
\begin{tabular}{@{}c|cccc|cccc@{}}
\toprule
\multirow{2}{*}{Methods} & \multicolumn{4}{c|}{2D tasks}                                                                                                                                                                                                                                      & \multicolumn{4}{c}{3D tasks}                                                                                                                                                                                                                                       \\
                         & \begin{tabular}[c]{@{}c@{}}Offline\\ training samples\end{tabular} & \begin{tabular}[c]{@{}c@{}}Success\\ rate\end{tabular} & \begin{tabular}[c]{@{}c@{}}Average\\ task error (cm)\end{tabular} & \begin{tabular}[c]{@{}c@{}}Average \\ task time (s)\end{tabular} & \begin{tabular}[c]{@{}c@{}}Offline\\ training samples\end{tabular} & \begin{tabular}[c]{@{}c@{}}Success\\ rate\end{tabular} & \begin{tabular}[c]{@{}c@{}}Average\\ task error (cm)\end{tabular} & \begin{tabular}[c]{@{}c@{}}Average \\ task time (s)\end{tabular} \\ \midrule
FKM+MPC                  & 180k                                                               & 82/100                                                 & 1.662                                                             & 10.488                                                            & 180k                                                               & 52/100                                                 & 3.298                                                             & 14.275                                                           \\
WLS                      & -                                                                  & 85/100                                                 & 0.992                                                             & 14.351                                                           & -                                                                  & 55/100                                                 & 1.940                                                             & 20.214                                                           \\
SAC                      & 1000k                                                              & 42/100                                                 & 3.185                                                             & 6.486                                                            & 1000k                                                              & 10/100                                                  & 3.412                                                             & 8.070                                                            \\
Ours(w/o online)         & \textbf{30k}                                                       & 94/100                                                 & 0.461                                                             & 8.568                                                            & \textbf{30k}                                                       & 69/100                                                 & 1.446                                                             & 9.521                                                            \\
Ours                     & \textbf{30k}                                                       & \textbf{97/100}                                        & \textbf{0.457}                                                    & 8.512                                                            & \textbf{30k}                                                       & \textbf{92/100}                                        & \textbf{1.254}                                                    & 9.529                                                            \\ \bottomrule
\end{tabular}
\vspace{-2mm}
\end{table*}

\begin{figure*} [tb]
  \centering 
  \subfigure[Case 1]{ 
    \includegraphics[width=2.8cm]{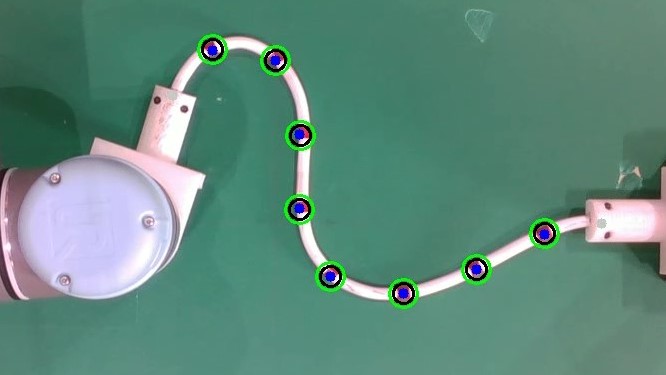} 
  } 
  \hspace{-0.4cm}
  \subfigure[Case 2]{ 
    \includegraphics[width=2.8cm]{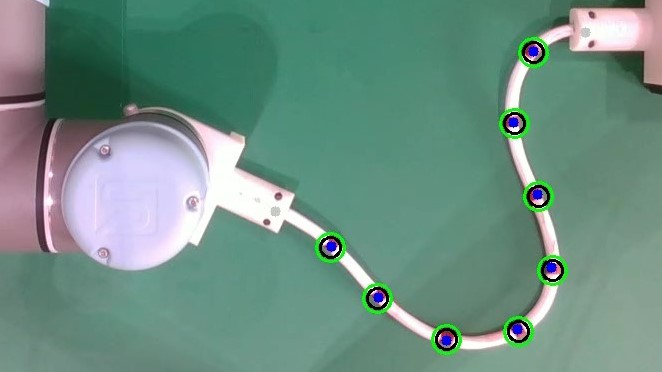} 
  }
  \hspace{-0.4cm}
  \subfigure[Case 3]{ 
    \includegraphics[width=2.8cm]{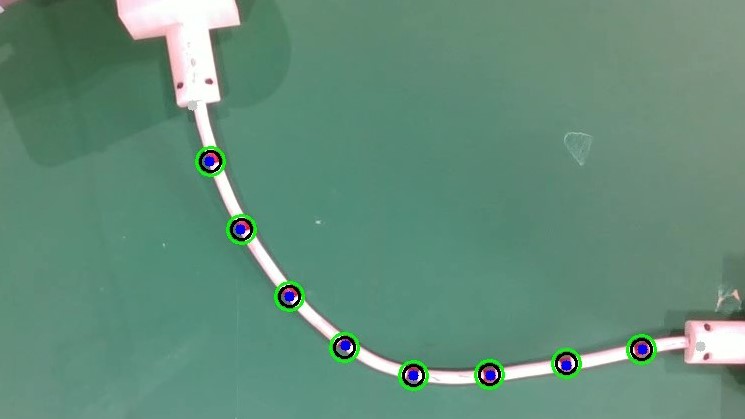} 
  }
  \hspace{-0.4cm}
  \subfigure[Case 6]{ 
    \includegraphics[width=2.8cm]{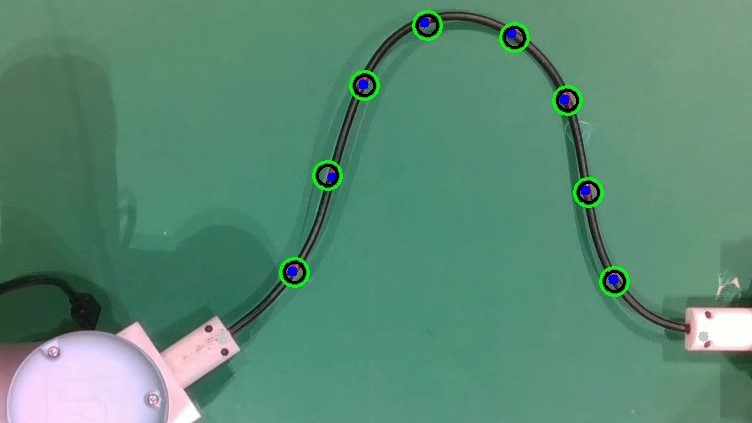} 
  }
  \hspace{-0.4cm}
  \subfigure[Case 7]{ 
    \includegraphics[width=2.8cm]{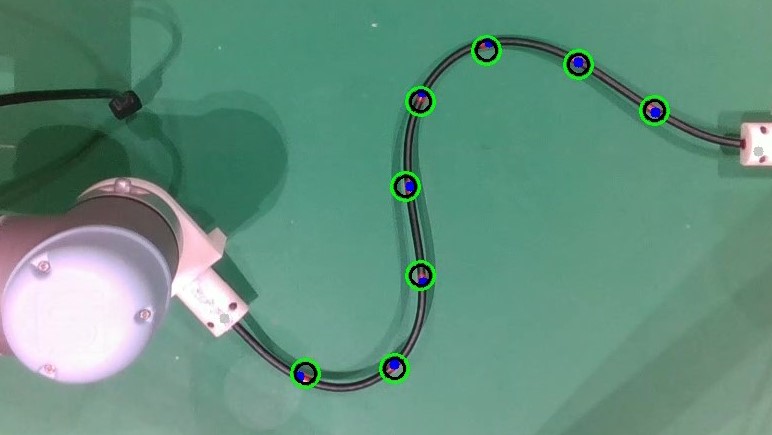} 
  }
  \hspace{-0.4cm}
  \subfigure[Case 8]{ 
    \includegraphics[width=2.8cm]{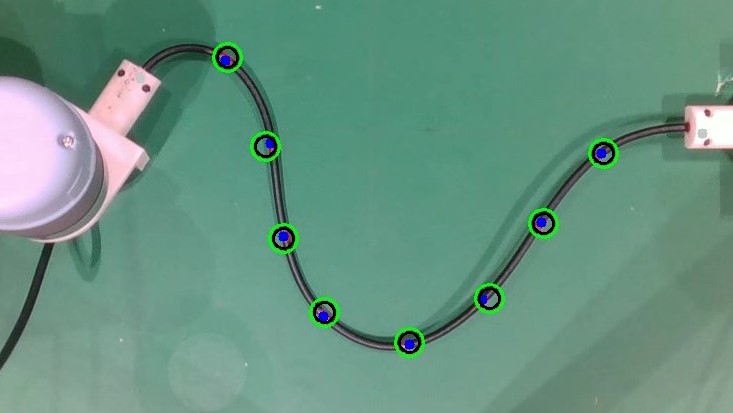} 
  }
  \vspace{-2mm}
  \caption{Some of the shape control tasks accomplished using our method in real-world experiments. The left end of the DLO is grasped by a UR5 arm and the right end is fixed. The green+black circles represent the desired positions of the DLO features. In all cases, the DLO starts from a straight line.}
  \label{fig:real_desired_shapes}
  \vspace{-4mm}
\end{figure*}

\begin{table}[tb]
\centering
\caption{Performance in real-world 2D shape control tasks.}
\vspace{-2mm}
\label{tab:real_control_results}
\begin{tabular}{@{}c|ccc@{}}
\toprule
Methods          & \begin{tabular}[c]{@{}c@{}}Success\\ rate\end{tabular} & \begin{tabular}[c]{@{}c@{}}Average\\ task error (cm)\end{tabular} & \begin{tabular}[c]{@{}c@{}}Average\\ task time (s)\end{tabular} \\ \midrule
FKM+MPC          & \textbf{10/10}                                         & 1.394                                                             & 7.670                                                           \\
WLS              & 9/10                                                   & 1.164                                                             & 19.089                                                          \\
SAC              & 1/10                                                   & 4.955                                                             & 12.300                                                          \\
Ours(w/o online) & \textbf{10/10}                                         & 1.153                                                             & 10.090                                                          \\
Ours             & \textbf{10/10}                                         & \textbf{0.620}                                                    & 8.220                                                           \\ \bottomrule
\end{tabular}
\vspace{-2mm}
\end{table}

\begin{figure} [tb]
  \centering 
  \subfigure[Case 3 (first DLO)]{ 
    \includegraphics[width=4.2cm]{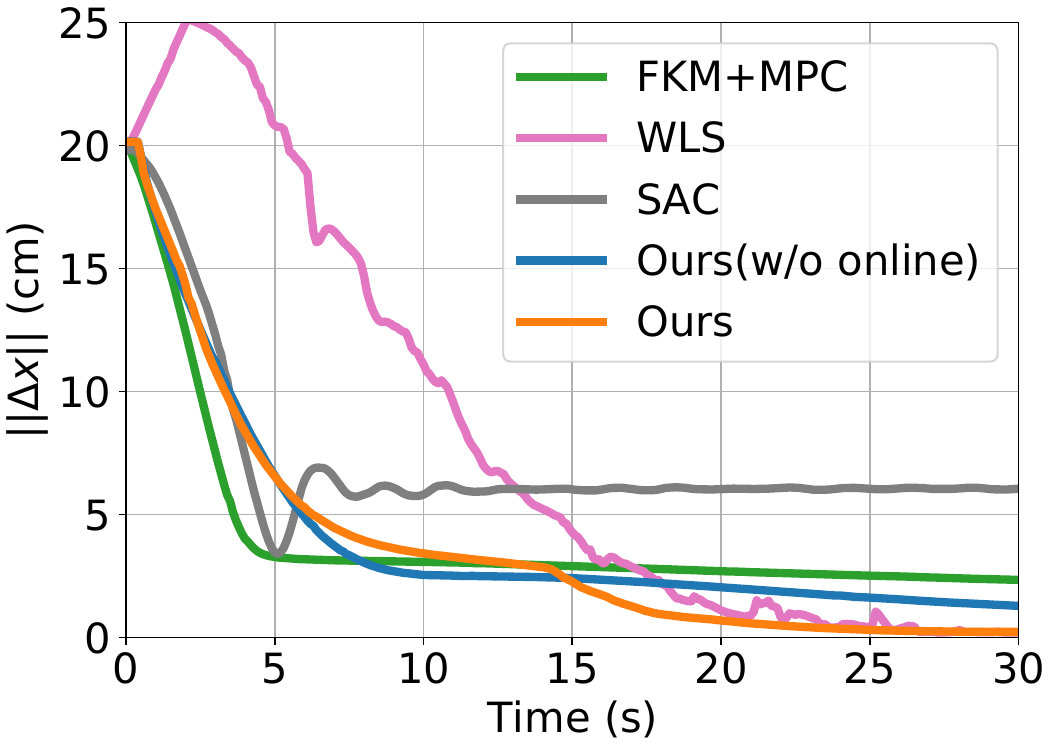} 
  } 
  \hspace{-0.5cm}
  \subfigure[Case 8 (second DLO)]{ 
    \includegraphics[width=4.2cm]{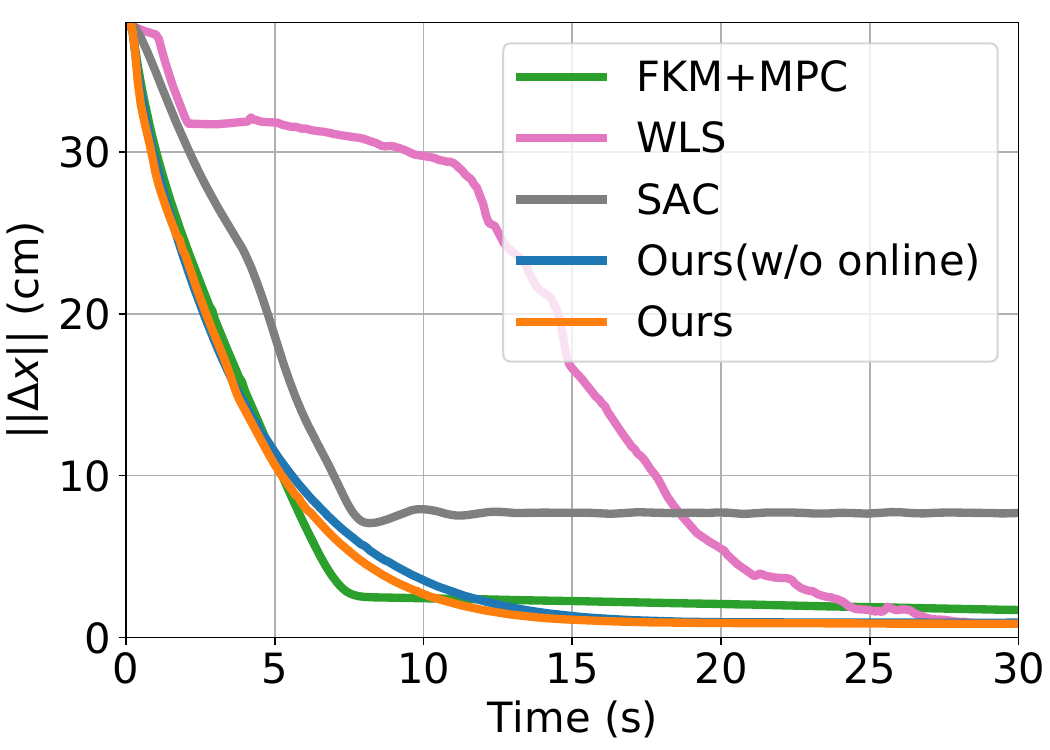} 
  }
  \caption{The control processes in real-world experiments. The $\| \Delta \bm x \|$ refers to the Euclidean distance between the current shape and the desired shape.}
  \label{fig:exp_real_control_plot}
  \vspace{-5mm}
\end{figure}

\subsection{Shape Control with Online Learning}

We evaluate the proposed method in DLO shape control tasks and compare it with other methods. All $m$ features are set as target points for shape control. During the tasks, the offline-trained models are used, and in our method the model will be further updated concurrently. In addition, domain randomization is applied during offline training to improve the models' generalization abilities. All the offline methods are trained on data collected on 10 DLOs with different lengths or diameters in simulation, while our method uses much less data than other offline methods. 

Several criteria are defined to evaluate the performance: (1)final task error: the Euclidean distance between the desired shape and final shape within 30s; (2)success rate: if the final task error is less than 5cm, this case is regarded successful; (3)average task error: the average of the final task errors over all successful cases; (4)average task time: the average time used to achieve success over all successful cases. Note that the task time is for reference only, since it depends on the control gain in servo methods or the sample range of control input in MPC or RL.

\subsubsection{Simulation}

First, we test their performance in both 2D and 3D DLO shape control tasks in simulation. The manipulated DLO is an untrained DLO, and 100 cases with different feasible desired shapes are tested. We also do the ablation study to validate the effect of the online learning in our method. The parameters are set as $\alpha=0.3, \eta_1=10^{-3}, \eta_2=50$. As shown in Table \ref{tab:sim_control_results}, our method significantly outperforms the compared methods on both success rate and average task error, even using much less offline training data. The task time of the online method WLS is the highest because it needs to initialize the Jacobian by moving the DLO ends in each DoF every time it starts. The average task error of FKM+MPC is higher because it has no further updating for the untrained manipulated DLO. The poor performance of SAC may be due to insufficient training. The contrast is starker in more challenging 3D tasks, where the success rates of compared methods are very low ($\leq$55\%), including our method without online learning (69\%), while our method with online learning achieves a 92\% success rate and the lowest average task error.

\subsubsection{Real-world experiments}

We also evaluate these methods in real-world 2D tasks. The same offline models as those in the simulation are used, which means no real-world data are collected for offline training. We separately carry out 5 tests with different feasible desired shapes on two DLOs: an electric wire with a length of 0.45m and diameter of 8mm, and an HDMI cable with a length of 0.6m and diameter of 5mm, as shown in Fig. \ref{fig:real_desired_shapes}. The parameters are set as $\alpha=0.3, \eta_1=10^{-3}, \eta_2=200$. The results are shown in Table \ref{tab:real_control_results}. Fig. \ref{fig:exp_real_control_plot} visualizes the control processes of two cases. Since the control input dimension is only $3$, all methods perform well in these relatively simple tasks except SAC. It is shown that the processes of WLS are slow and unsmooth, and those of FKM+MPC are fast but less precise. Our method completes all 10 tasks and achieves the lowest average task error, where the online learning enables faster and more precise control.

\section{Conclusion}

This paper considers the shape control of DLOs with unknown deformation models. We formulate the deformation model in a local linear format, which is estimated in both offline and online phases. First, the offline learning well initiates the estimation of the model. Then, the adaptive control scheme with online learning further updates the model and achieves the shape control tasks in the presence of an inaccurate offline model. The experiments demonstrate that the offline learning of the local linear deformation model is accurate and data-efficient. By combining the offline and online learning, our method outperforms the compared methods, and adapts well to untrained DLOs. Future work will include more detailed analysis of our method and validations in real-world 3D dual-arm manipulation tasks using a high-precision 3D camera.









\newpage

\bibliographystyle{IEEEtran}

\end{document}